# Sesame Plant Segmentation Dataset: A YOLO Formatted Annotated Dataset

Sunusi Ibrahim Muhammad
*Department of Petroleum Engineering*
*Bayero University Kano*
Kano, Nigeria
sunusimuhammada@gmail.com

Ismail Ismail Tijjani
*Department of mechatronics Engineering*
*Bayero University Kano*
Kano, Nigeria
ismailismailtj@gmail.com

Saadatu Yusuf Jumare
*Department of Plant Science*
*Ahmadu Bello University zaria*
Kaduna, Nigeria
saadatulanaamjumare@gmail.com

Fatima Isah Jibrin
*Department of Computer Science*
*Gombe State University*
Gombe, Nigeria
fateemaij@gmail.com

*Abstract*— **This paper presents the Sesame Plant Segmentation Dataset, an open-source annotated image dataset contributed to support the development of Artificial Intelligence (AI) models for agriculture, with a specific focus on sesame plants. The dataset Comprising 206 training images, 43 validation images and 43 for testing in YOLO compatible segmentation format, the dataset captures sesame plants at early growth stages under varying environmental conditions. Data were collected using a high resolution mobile camera from farms in Jirdede, Daura Local Government Area, Katsina State, Nigeria, and annotated using the Segment Anything Model (SAM-2) model on the Roboflow platform with farmer supervision. Unlike traditional bounding box datasets, this dataset employs precise pixel level segmentation, enabling robust detection and analysis of sesame plants in real-world farm settings. The model was evaluated using the Ultralytics YOLOv8 framework, yielding strong performance across both bounding box and segmentation tasks. For bounding box detection, the model achieved a Recall of 79%, Precision of 79%, mean Average Precision at IoU 0.50 (mAP@50) of 84%, and mAP@50–95 of 58%. In the segmentation task, it recorded a Recall of 82%, Precision of 77%, mAP@50 of 84%, and mAP@50–95 of 52. the dataset represents a *novel contribution* to sesame-focused agricultural vision datasets locally in Nigeria. Potential applications include real-time plant monitoring, yield estimation, and agricultural research. The dataset is publicly available on Kaggle at**

https://www.kaggle.com/datasets/ismailismailtijjani/sesame-plant-detection-dataset

*Keywords—Sesame Plant, Instance Segmentation, YOLOv8, Precision Agriculture, Computer Vision, Northern Nigeria, Open Source Dataset*

I. INTRODUCTION

Nigeria is blessed with abundant natural resources, including 70.8 million hectares of agricultural land across agro-ecological zones that can support a wide range of food products [1]. The strategic importance of the agricultural sector to the Nigerian economy cannot be overemphasized, as its contribution to GDP hovered between 24.45% in 2016 and 25.70% in 2020 [1] [2]. Among the valuable crops with high economic potential is Sesame. Sesame cultivation in Nigeria has expanded significantly since the early 2000s, transitioning from a subsistence crop to a major cash crop. Its growth has been driven by increasing international demand, government support for non-oil exports, and the crop's adaptability to arid conditions [3]. Sesame is widely grown as a cash crop within the savanna agro-ecological zones, particularly in the central and northern parts of the Nigeria, such as Kano, Jigawa, Benue, Gombe, Plateau, Kaduna, Nasarawa, Katsina, and Borno States, as well as the Federal Capital Territory [4]. Sesame can be cultivated on less fertile land where other crops may struggle to grow and develop properly [5]. It has also been found to thrive on marginal lands and rough terrains due to its resistance to drought and high temperatures. Thus, it is one of the most resilient crops that can grow effectively in the arid climate of the West African Sahel [6]. Despite these advantages, many farmers face challenges such as the inability to monitor crops for early disease and pest infestations, which cause widespread damage before they can be addressed, difficulty in optimizing irrigation, manual assessment of crop health and yield, and inefficient harvesting methods [7].

The rapid advancement of AI applications in agriculture has begun to address many of these challenges [7]. However, these technologies require large amounts of high-quality locally sourced data. This underscores the need for high-quality annotated datasets to develop robust and accurate models [7][8]. Such models are important for real-time crop monitoring, providing efficient solutions that conserve resources and time while enhancing agricultural productivity [8]. Most existing agricultural datasets are derived from diverse geographical regions and often fail to reflect the unique environmental features of local contexts, such as variations in climate, soil types, plant health conditions, and the specific characteristics of regional crop varieties, these disparities hinder the development of regionally effective AI solutions [9][10]. Traditional plant detection models, which rely on bounding box annotations, also fall short in scenarios involving dense foliage or overlapping leaves, limiting their precision in complex farm environments [11]. To address these challenges, we introduce the Sesame Plant Segmentation Dataset, a locally curated and annotated collection of images formatted for YOLO segmentation tasks. This dataset enables precise pixel-level localization and analysis of sesame plants, overcoming the limitations posed by intricate farm settings. The dataset was collected in Jirdede, Daura Local Government Area of Katsina State, Nigeria. It is designed to empower a wide range of stakeholders including AI developers, agricultural engineers, domain experts, business owners, entrepreneurs, policymakers, and agricultural extension officers. It supports the creation of prototypes, proof-of-concept studies, and scalable applications aimed at improving crop management.

This paper outlines the dataset structure, the methodology for data collection and annotation, its potential use cases, and plans for future expansion positioning it as a foundational resource for advancing precision agriculture in Nigeria and beyond.

## II. RELATED WORK

In recent years, significant efforts have been made by researchers and organizations to curate and open-source datasets that advance AI and machine learning research. Initially, most of these datasets were tabular or structured in nature. One of the earliest and most influential is the Iris dataset, introduced in 1936 [12], which became a standard benchmark for classification problems. This was followed by several widely used datasets such as the Wine Quality dataset [13], Titanic dataset [14], Breast Cancer Wisconsin dataset [15], and the Bank Marketing dataset [16]. These benchmarks have contributed significantly to the development and evaluation of supervised learning models, particularly for classification and regression tasks.

With the growth of computer vision, the focus shifted toward image-based datasets. The MNIST dataset, introduced in 1998 [17], consists of 70,000 grayscale images of handwritten digits and quickly became a foundational benchmark for image classification models. Despite its simplicity, it played an important role in early deep learning experiments. However, its limited complexity led to the development of more advanced datasets such as CIFAR-10 and CIFAR-100 [18], which introduced low-resolution colored images across 10 and 100 classes, respectively. These datasets became standard for training convolutional neural networks (CNNs). With continues advancement in computer vision model go beyond classification to object detection and localization which rises the need for annotated dataset, to provide richer context and more complex annotation types, datasets like COCO (Common Objects in Context) [19] and PASCAL VOC [20] were introduced. COCO, with over 2.5 million labeled instances, supports tasks such as object detection, segmentation, and image captioning. PASCAL VOC focuses on real-world object recognition and segmentation in diverse settings. Other large-scale datasets containing millions to billions of open-source dataset have further contributed to advancements in AI research [21][22][23]. These datasets have become resourceful for evaluating and understanding model performance in complex visual environments.

In the agricultural domain, the need for specialized datasets has also grown. The Food-101 dataset [24], which includes 101,000 images across 101 food categories, has been widely used for food recognition and dietary monitoring. Additional datasets [25][26][27] have targeted various food-related image recognition tasks, including processed food detection.

For crop health monitoring, the PlantVillage dataset [28], introduced in 2015, remains one of the most important contributions. It contains over 54,000 images of healthy and diseased plant leaves from various crop species and has become a primary resource for plant disease classification. Similarly, the DeepWeeds dataset [29] includes field images of eight weed species and has been instrumental in developing weed detection models for smart farming systems. The PlantDoc dataset [30], which includes 2,598 images covering 13 plant species and 17 disease categories, and AgriSeg [31], designed to support AI-driven robotic systems for precision agriculture using both synthetic and real-world image data, have further expanded the field. In recent years, sesame-related datasets have also been introduced. The Sesame Disease Dataset, released on Roboflow [32], focuses on disease detection, while the Sesame-Weed Aerial Dataset [33] supports research in agricultural precision and weed identification from aerial imagery.

Despite these advancements, significant limitations remain. Most existing datasets are developed in foreign environments and are not representative of local agricultural systems, particularly in Africa. This limits their applicability in real-world deployment within regional farming contexts.

Although several datasets have been introduced related to sesame, there is no existing open-source dataset that specifically targets sesame plants in Nigeria or provides segmentation annotations in YOLO format. the Sesame Plant Segmentation Dataset, addresses this gap. It is the first localized dataset focused on a Nigerian Sesame crop, annotated specifically for real-time segmentation tasks. This makes it a valuable resource for advancing research in agricultural AI, particularly in underrepresented regions.

III. METHODOLOGY

The Sesame Plant Segmentation Dataset was created through a structured pipeline involving data collection, Image preparation, annotation, and formatting for YOLO-based segmentation models. The goal was to capture real-world sesame farming conditions and provide detailed, pixel-level segmentation annotations suitable for training computer vision Model.

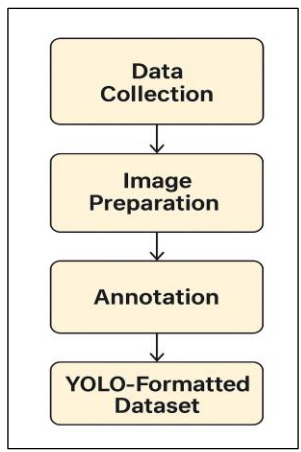

Fig. 1. Methodology Pipeline

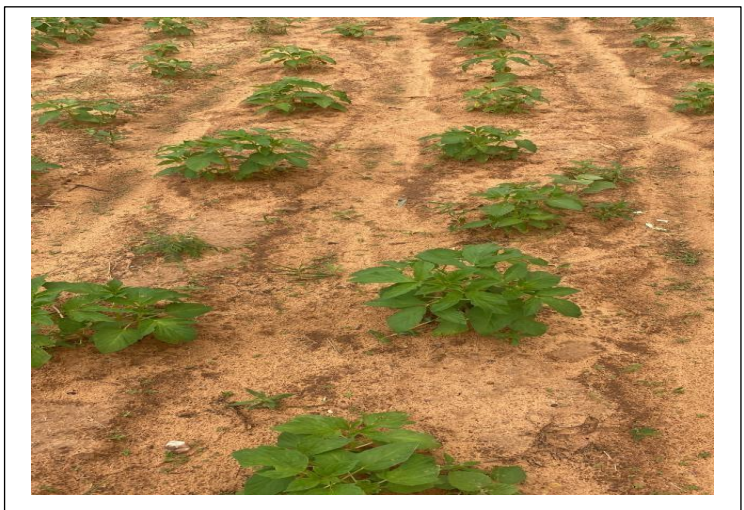

Fig. 2. Sesame plant captured at its early growth stage in Jirdede, Nigeria

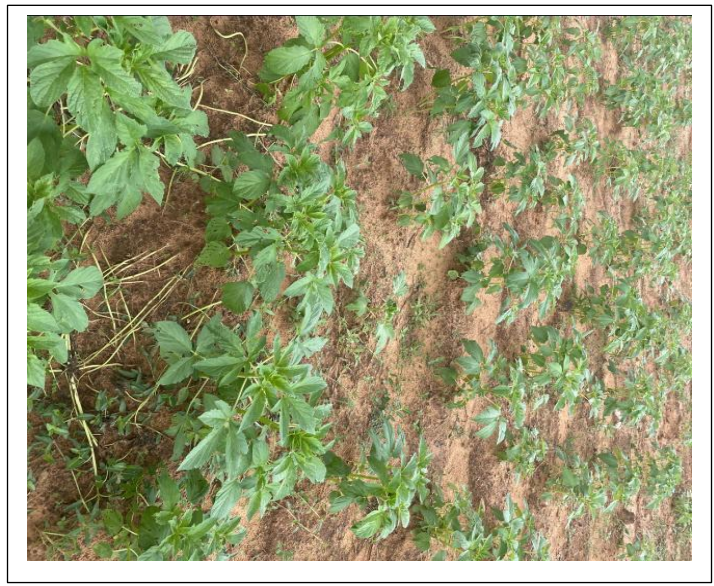

Fig. 3. Sesame plant captured at its early growth stage in Jirdede, Nigeria

*A. Data Collection*

The first and most important step in any machine learning project is data collection. Images were manually collected from sesame farms located in Jirdede, Daura Local Government Area, Katsina State, Nigeria, using an iPhone 11 camera, the camera has both wide ($f$/1.8) and ultra-wide ($f$/2.4) 12-megapixel lenses. all images were taken under natural daylight during the rainy season, as shown in Figure 2 and 3. The Images were captured from various perspectives including horizontal, top, and side views to ensure that all orientations of the plant were represented. This approach was adopted to ensure that the dataset reflected authentic and variable farm environments typical of Northern Nigeria. The dataset primarily focuses on sesame plants at early growth stages (45–85 days) during seed development. During collection, approximately 95% of the plants were healthy and unaffected by disease, allowing for clear, unobstructed images of the sesame plants, Capturing plants at various developmental stages enables models trained on this dataset to recognize differences in leaf shape, overlap, and density features that are important for accurate plant segmentation, especially in dense or complex farm scenes [34].

*B. Image Preparation*

after data collection, the images were prepared for annotation. Each image was cropped and resized to a uniform resolution of 640×640 pixels. This preprocessing ensured consistency across the dataset and improved model performance by standardizing input dimensions. Maintaining uniform image size helped preserve fine structural details of the sesame plants, which are essential for high-quality instance segmentation [35].

*C. Annotation*

Annotation is a fundamental stage in computer vision dataset development. It involves labeling images or marking specific regions to create meaningful data from which models can learn to recognize and localize objects within image . For any object detection or segmentation task, annotation represents the ground truth that guides the learning process during model training. There are numerous tools available for image annotation, both offline and online for computer vision application [36]. Offline tools include LabelMe, LabelImg, and CVAT, while popular online platforms include Roboflow Annotate, V7 Labs, MakeSense.ai, and PixLab Annotate. Each offers unique features depending on project requirements from simplicity and accessibility to automation and collaborative annotation. Annotations can take several forms depending on the computer vision task:

- Bounding Box: Rectangular boxes drawn around objects to localize them within an image. However, this method may not capture the complex shapes of plant structures [37].
- Polygonal Segmentation: Outlines objects using connected vertices to form polygons, allowing for precise representation of irregular and detailed plant boundaries that bounding boxes cannot capture [38].
- Semantic Segmentation: Assigns each pixel a class label, grouping all pixels belonging to the same category. However, it does not distinguish between individual instances of objects within the image [39].
- Instance Segmentation: Extends semantic segmentation by assigning class labels to each pixel while also distinguishing between individual object instances. This method is particularly effective for identifying overlapping plants or those with morphological variations [40].

for this dataset, instance segmentation was chosen to account for variations in plant structure, overlap, and health conditions. This approach enables models to learn fine-grained segmentation, which is especially important in real-world farming environments where plants often overlap or grow densely. The annotation process was performed using SAM-2[23] integrated on the Roboflow platform, under the supervision of local farmers serving as domain experts. Their involvement ensured annotation accuracy and real-world validity, resulting in high-quality labeled data that accurately represents natural sesame farm conditions.

### D. YOLO formatted Dataset

After annotation, the dataset was exported from Roboflow in YOLO segmentation format. The YOLO (You Only Look Once) framework was chosen due to its speed, flexibility, and strong real-time performance making it a leading choice for modern object detection tasks[41][42]. YOLO processes each image once, simultaneously detecting and classifying objects[40]. This single-step detection approach enables fast and accurate results ideal for applications such as automated agriculture, drone-based crop monitoring, and precision farming where quick, reliable detection is essential[41]. By adopting the YOLO format, the dataset becomes ready to use for state of the art computer vision models, balancing efficiency, accuracy, and real-time performance across a range of hardware configurations.

The final dataset consists of **293 annotated images**, organized as follows:

- 206 images (70**%)** for training
- 43 images (15**%)** for validation
- 43 images (15%) for testing

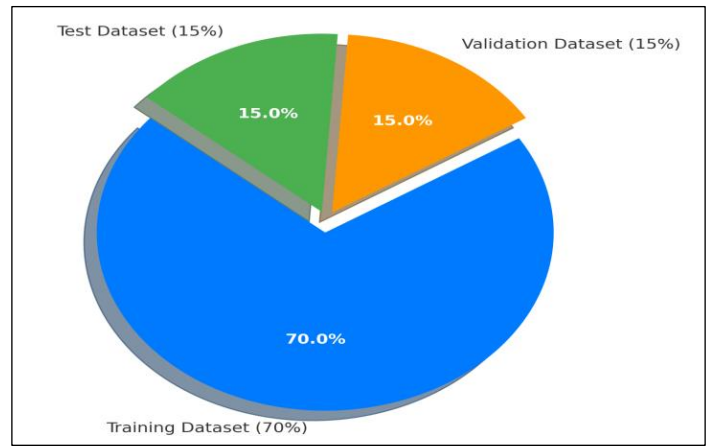

Fig. 4. Dataset Split Distribution for Model Development

The dataset is organized in a simple directory structure:

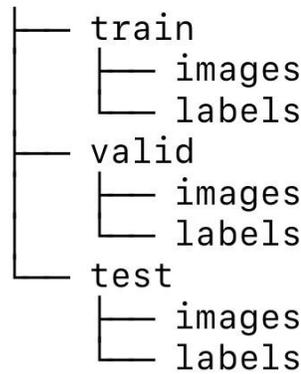

## IV. DATASET EVALUATION

To evaluate the performance and practical applicability of the Sesame Plant Segmentation Dataset, a model was trained using Ultralytics YOLOv8, a state of the art framework for real-time object detection and instance segmentation. The decision to use YOLOv8 was driven by its performance and advanced architecture at the time of the dataset's release. As the most recent and optimized version in the YOLO family, YOLOv8 introduced significant improvements in accuracy, speed, and segmentation quality, making it a good choice for benchmarking datasets formatted in the YOLO structure [42].

The model was trained to evaluate how well the dataset supports accurate segmentation of sesame plants in realistic field scenarios, including conditions such as overlapping leaves, dense vegetation, and varying natural lighting. The training was conducted using YOLOv8's default configuration to provide a fair and consistent benchmark. the model was trained for 100 epochs with a batch size of 16 and an image resolution of 640×640 pixels, using the optimizer set to 'auto' to dynamically adjust learning during the process. A total of 70% of dataset were used for training, 15% for validation, and 15% for testing. This setup provided a foundation for evaluating the dataset's capability in supporting effective segmentation of sesame plants under diverse real-world conditions.

## V. RESULTS

The model exhibited strong segmentation capabilities, effectively delineating the contours of sesame plants even under challenging field conditions. By employing instance segmentation rather than traditional bounding-box detection, the model achieved superior spatial precision and fine-grained localization, allowing it to distinguish individual leaves and stems within dense plant clusters. This level of detail is particularly important for agricultural applications where precise plant boundary detection enhances downstream tasks such as growth analysis, disease monitoring, and yield estimation.

Model was quantitatively assessed using standard evaluation metrics, including Precision (P), Recall (R), and mean Average Precision (mAP) at Intersection over Union (IoU) thresholds of 0.50 and across a range of 0.50–0.95. These metrics provide a comprehensive view of the model's ability to detect, localize, and segment sesame plants accurately. The results are summarized in Table 1, where both the bounding-box and segmentation mask performances are reported. The model achieved a Precision of 79% for bounding boxes and 82% for segmentation masks, indicating that the segmentation approach produced slightly more accurate predictions. Similarly, the Recall values were 79% and 77%, respectively, demonstrating a balanced detection ability across both methods. The mAP@50 scores were 84% for both bounding-box and segmentation outputs, reflecting strong detection accuracy, while the more stringent mAP@50–95 metric yielded 58% for bounding boxes and 52% for segmentation masks, underscoring the model's robustness under varying IoU thresholds.

These results confirm that the Sesame Plant Segmentation Dataset is highly effective for training segmentation models aimed at precision agriculture. The visual outputs further support this conclusion. As illustrated in Figure 6, the model successfully identified sesame plants using bounding-box predictions, while Figure 7 presents the corresponding segmentation masks with color overlays that highlight the model's fine-grained detection capability. Together, these visual and quantitative results demonstrate that the dataset enables high-performance modeling in complex agricultural environments.

These evaluation establishes the Sesame Plant Segmentation Dataset as a valuable resource for developing intelligent vision-based agricultural systems. Its ability to facilitate accurate segmentation under realistic field conditions makes it particularly relevant for real-time crop monitoring, smart spraying systems, yield estimation, and agricultural automation, especially in regions with environmental conditions similar to Northern Nigeria making the dataset openly accessible, it was aim to empower AI researchers, agricultural engineers, startups, and institutions to build intelligent systems that are locally relevant, scalable, and practical.

TABLE I. EVALUATION METRICS FOR YOLOV8 MODEL TRAINED ON SESAME PLANT SEGMENTATION DATASET

| Metrics | Bounding Box Metrics | Segmentation Mask Metrics |
|---|---|---|
| Precision (P) | 79% | 82% |
| Recall (R) | 79% | 77% |
| mAP@50 | 84% | 84% |
| Map@50-90 | 58% | 52% |

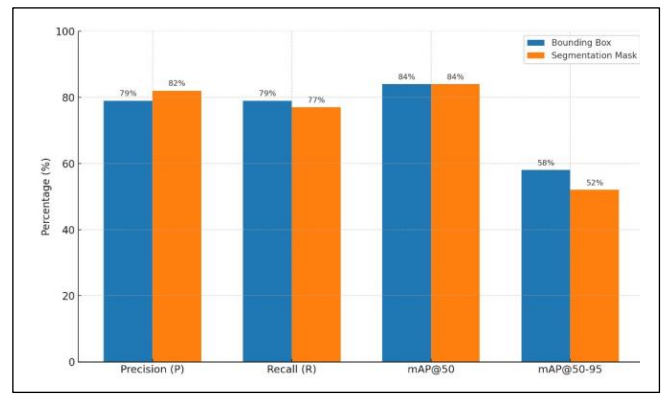

Fig. 5. Performance Metrics for Bounding Box and Segmentation Mask

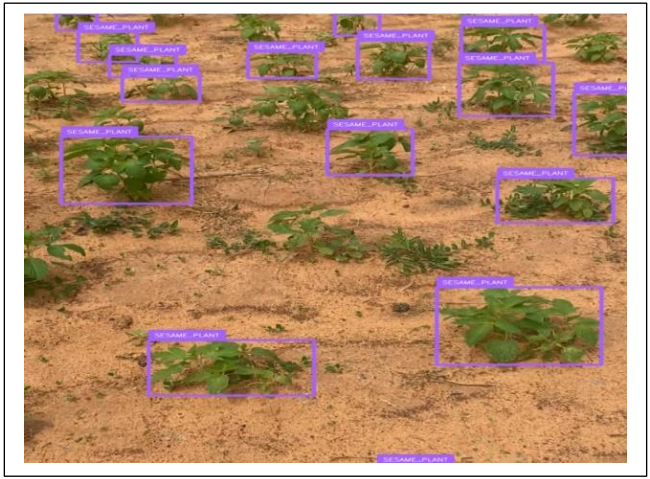

Fig. 6. Predicted output of sesame Plants by bounding Box

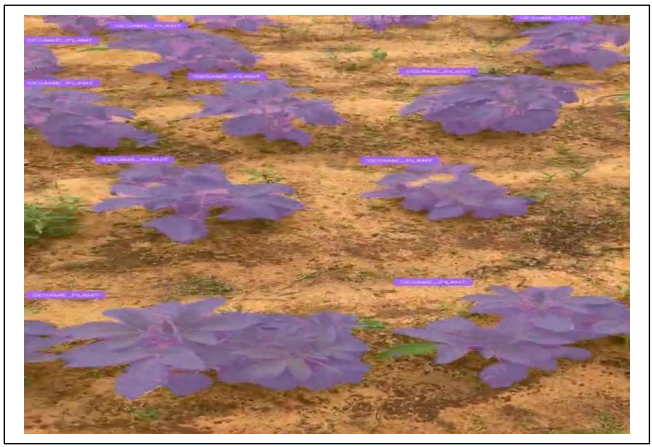

Fig. 5. Predicted output of sesame plants Segmentation.

## VII. LIMITATION AND FUTURE WORK

While the current dataset provides a foundation, it has some limitations that plan to address in future versions. These include:

- Expanding the dataset to cover more growth stages of sesame plants.
- Capturing images under different lighting conditions, including dawn, dusk, and overcast skies, to improve model robustness.
- Including additional annotation classes such as weeds, diseased plants, and other background elements, which will allow for multi-class segmentation and more advanced research.
- Increasing dataset size and diversity by collecting from more farms across different regions to improve generalizability.

These improvements will make the dataset even more valuable for real-world deployment and advanced machine learning research in agriculture

## VI. CONCLUSION

In this research, we introduced the Sesame Plant Segmentation Dataset, a locally sourced and annotated image dataset designed to support AI applications in agriculture, particularly in Northern Nigeria. By focusing on sesame, high value yet underrepresented crops, we address a critical gap in the availability of open-source agricultural datasets tailored to local environmental conditions.

The dataset includes pixel-level instance segmentation annotations formatted for YOLO-compatible models, making it especially suitable for real-time detection and precision agriculture applications. Through evaluation using the YOLOv8 architecture, demonstrated the dataset's strong performance in segmenting sesame plants with high precision and recall, even in complex field scenarios. This shows its potential for use in tasks such as crop monitoring, smart spraying, weed detection, and automated yield estimation, by

## ACKNOWLEDGMENT

We express gratitude to the farmers in Jirdede, for their support and cooperation during data collection and annotation. We also acknowledge the Roboflow platform for facilitating efficient annotation.